\pdfoutput=1
\documentclass[11pt]{article}

\usepackage[preprint]{acl}

\usepackage{times}
\usepackage{latexsym}

\usepackage[T1]{fontenc}
\usepackage[utf8]{inputenc}
\usepackage{amsmath}
\usepackage{amssymb}
\usepackage{amsfonts}
\usepackage{bm}
\usepackage{algorithm}
\usepackage{algorithmicx}
\usepackage{algpseudocode}
\usepackage{wrapfig}

\usepackage{bbding}

\usepackage{amsmath}
\usepackage{amssymb}
\usepackage{booktabs}      % for well‐spaced tables if needed
\usepackage{multirow}      % for multirow cells in tables
\usepackage{xcolor}        % for row coloring in tables
\usepackage{graphicx}      % for figures
\usepackage{caption}       % for caption control
\usepackage{subcaption}    % if you need subfigures

\usepackage{microtype}

\usepackage{inconsolata}

\usepackage[T1]{fontenc}
\usepackage{colortbl}
\usepackage{booktabs,multirow}
\usepackage{makecell}
\usepackage{tabulary}
\usepackage{fontawesome5}
\usepackage{pifont}
\usepackage{tikz}

\newlength\savewidth\newcommand\shline{\noalign{\global\savewidth\arrayrulewidth
  \global\arrayrulewidth 1pt}\hline\noalign{\global\arrayrulewidth\savewidth}}

\newcommand{\tablestyle}[2]{\setlength{\tabcolsep}{#1}\renewcommand{\arraystretch}{#2}\centering\footnotesize}

\newcommand{\cmark}{\ding{51}}%
\title{Video Compression Commander: \\ Plug-and-Play Inference Acceleration for Video Large Language Models}

\author{Xuyang Liu$^{1,2}$\thanks{Equal contribution.  Work done during a visit to the EPIC Lab at Shanghai Jiao Tong University. \\ \text{\Envelope} Corresponding author: zhanglinfeng@sjtu.edu.cn} \quad Yiyu Wang$^{1*}$ \quad Junpeng Ma$^{3}$ \quad Linfeng Zhang$^{1}$$^{\text{\Envelope}}$ \\
$^{1}$Shanghai Jiao Tong University \quad
$^{2}$Sichuan University \quad
$^{3}$Fudan University \\
}
\begin{document}
\maketitle

\begin{abstract}

Video large language models (VideoLLM) excel at video understanding, but face efficiency challenges due to the quadratic complexity of abundant visual tokens. 
Our systematic analysis of token compression methods for VideoLLMs reveals two critical issues: \textbf{(i)} overlooking distinctive visual signals across frames, leading to information loss; \textbf{(ii)} suffering from implementation constraints, causing incompatibility with modern architectures or efficient operators.
To address these challenges, we distill three design principles for VideoLLM token compression and propose a plug-and-play inference acceleration framework ``\textbf{Vid}eo \textbf{Com}pression \textbf{Com}mander'' (\textbf{VidCom$^2$}). 
By quantifying each frame’s uniqueness, VidCom$^2$ adaptively adjusts compression intensity across frames, effectively preserving essential information while reducing redundancy in video sequences. 
Extensive experiments across various VideoLLMs and benchmarks demonstrate the superior performance and efficiency of our VidCom$^2$. 
With only \textbf{25\%} visual tokens, VidCom$^2$ achieves \textbf{99.6\%} of the original performance on LLaVA-OV while reducing \textbf{70.8\%} of the LLM generation latency. 
Notably, our Frame Compression Adjustment strategy is compatible with other token compression methods to further improve their performance. Our code is available at \url{https://github.com/xuyang-liu16/VidCom2}.
\end{abstract}

\section{Introduction}

\begin{figure}[t]
    \centering
    \includegraphics[width=\linewidth]{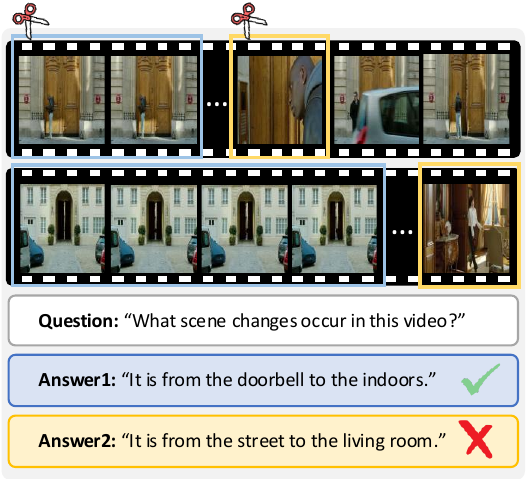}
    \vspace{-7mm}
    \caption{\textbf{Power of frame uniqueness.} Removing \textcolor[RGB]{56,94,181}{\textbf{24} redundant frames} results in \textcolor[RGB]{29,195,136}{\textbf{accurate}} video understanding by VideoLLMs, while dropping just \textcolor[RGB]{248,170,52}{\textbf{8} unique frames} leads to \textcolor[RGB]{223,52,39}{\textbf{inaccurate}} video comprehension, highlighting the critical role of unique frames for VideoLLMs.}
    \vspace{-5mm}
    \label{fig:motivation}
\end{figure}

Recently, Video Large Language Models (VideoLLMs) have demonstrated remarkable performance in video understanding and reasoning tasks~\cite{Zhang:Video-LLaMA,wang2025internvideo2.5}. However, videos inherently contain multiple consecutive frames, resulting in a significantly higher number of visual tokens compared to images. For instance, LLaVA-OneVision~\cite{li2024llava-ov} processes $32 \times 196$ visual tokens per video, while LLaVA-Video~\cite{zhang2024llava-video} handles even more at $64 \times 182$ visual tokens. This high token count inevitably leads to expensive computation~\cite{liu2025shifting}, especially for long video understanding~\cite{chen2024longvila}.

To mitigate this computational burden, researchers have turned to token compression methods~\cite{Chen:FastV,Yang2024:Visionzip}, considering the inherent visual redundancy and aiming to minimize redundant visual information. 
These approaches can be categorized as pre-LLM~\cite{Zhang:FasterVLM} or intra-LLM~\cite{Chen:FastV} methods, based on whether compression occurs before or within the LLM.
Most of these methods are training-free, enabling plug-and-play inference acceleration for existing VideoLLMs. 
However, despite these efforts, existing token compression methods suffer from \textbf{\emph{two critical issues}}: 

% \paragraph{(I) Design Myopia}
\noindent \textbf{(I) Design Myopia:} In human video perception, we naturally focus on distinctive frames (\textit{e.g.}, those with significant spatio-temporal changes) while ignoring repetitive and redundant visual information~\cite{ma2025mmg-vid}. By contrast, most existing token compression methods apply a \textit{uniform} compression strategy across all frames, treating each one as equally informative. Even recent VideoLLM-specific method DyCoke~\cite{tao2024dycoke} exhibits this limitation by grouping every four consecutive frames into a fixed window and compressing them identically, without regard for the varying distinctiveness of individual frames. Figure~\ref{fig:motivation} further illustrates the critical nature of this issue: removing 24 redundant frames does not affect the accurate response of the LLaVA-OneVision, whereas dropping just 8 unique frames causes it to fail, despite being only a \textbf{third} of the number. This contrast shows that uniform compression risks discarding critical information in unique frames that VideoLLMs may rely on, thereby significantly impacting overall performance. Notably, Table~\ref{tab:main_results} indicates that some methods even \textbf{underperform} random token dropping, further indicating their sub-optimal performance.

\begin{table}[!t]
  \centering
  \setlength{\tabcolsep}{0.4pt} % 调整列间距
    \scalebox{0.73}{\begin{tabular}{lcccccc}
    \multirow{2}{*}{\textbf{Methods}} & \textbf{Pre-} & \textbf{Intra-} & \textbf{\texttt{[CLS]}} & \textbf{Video-} & \textbf{Frame} & \textbf{Efficient} \\
    
    & \textbf{LLM} & \textbf{LLM} & \textbf{Dependency} & \textbf{Specific} & \textbf{Uniqueness} & \textbf{Attention} \\
    \shline
     FastV &  & \cmark &  &  &  &  \\
     PDrop &  & \cmark &  &  &  &  \\
     SparseVLM &  & \cmark &  &  &  & \\
     MUSTDrop & \cmark & \cmark & \cmark &  &  &  \\
     FiCoCo & \cmark & \cmark & \cmark &  &  &  \\
     FasterVLM & \cmark &  & \cmark &  &  & \cmark \\
     DyCoke & \cmark &  &  & \cmark &   & \cmark \\
    \rowcolor[rgb]{ .949,  .949,  .949}
    \textbf{VidCom$^2$} & \cmark &  &  & \cmark &  \cmark & \cmark \\
    \end{tabular}%
    }
    \vspace{-2mm}
  \caption{\textbf{Feature comparison with existing training-free token compression methods.} Most suffer from design myopia and implementation constraints.}
  \label{tab:para_comparsion}%
  \vspace{-5mm}
\end{table}%

% \paragraph{(II) Implementation Constraints}
\noindent \textbf{(II) Implementation Constraints:} 
Beyond design limitations, existing methods face practical constraints. Some token compression works~\cite{Zhang:FasterVLM,Liu2024:MUSTDrop} rely on \texttt{[CLS]} attention weights in ViT for informative token preservation, yet modern VideoLLMs adopt SigLIP~\cite{ZhaiM0B23:SigLIP} as visual encoder without \texttt{[CLS]} token. Meanwhile, certain methods~\cite{Zhang:SparseVLM,xing2024pdrop} aim to leverage textual information but require explicit attention weights in specific LLM layers, making them incompatible with efficient attention operators~\cite{Dao2022:FlashAttention}. This incompatibility leads to higher peak memory usage, even \textbf{surpassing} that of uncompressed processing (see Table~\ref{tab:efficiency_comparisons}), which is especially problematic for long video understanding~\cite{wen2025token,wen2025dart}.

We summarize existing works in Table~\ref{tab:para_comparsion} and identify \textbf{\emph{three key principles}} for designing effective and efficient token compression methods for VideoLLM:
\textbf{(i) Model Adaptability:} The method should be easily compatible with and adaptable to the majority of existing VideoLLMs~\cite{zhang2024llava-video,Wang:Qwen2-VL}; \textbf{(ii) Frame Uniqueness:} The method should consider varying distinctiveness across video frames;
\textbf{(iii) Operator Compatibility:} The method should maintain compatibility with efficient operators~\cite{daoFlashAttention-2}.

Based on above analysis, we propose ``\textbf{Vid}eo \textbf{Com}pression \textbf{Com}mander'' (\textit{i.e.}, \textbf{VidCom$^2$}), an efficient plug-and-play token compression method for VideoLLMs from the perspective of frame uniqueness. Our VidCom$^2$ follows a principled two-stage approach: first adjusting frame-wise compression intensity based on each frame's uniqueness in the video sequence, then performing token compression by evaluating token distinctiveness both within individual frames and across the entire video. Through this careful design, VidCom$^2$ mimics human video perception by adaptively adjusting attention to different frames (see Figure~\ref{fig:frame_uniqueness_vidcom2}), preserving information from key frames while minimizing redundant visual content.

In summary, our contributions are three-fold:
\begin{itemize}

    \item \textbf{Empirical Method Analysis:} We critically analyze existing token compression methods, unveiling their inherent limitations and delineating three key design principles for effective and efficient VideoLLM token compression.
    
    \item \textbf{Video Compression Commander:} We are the first to propose a VideoLLM token compression framework based on frame uniqueness, offering a plug-and-play method with frame-wise dynamic compression.
    
    \item \textbf{Outstanding Performance \& Efficiency:} Extensive experiments on diverse benchmarks demonstrate superior efficiency-performance trade-offs. With 15\% tokens, VidCom$^2$ outperforms the second-best method by \textbf{3.9\%} and \textbf{2.2\%} on LLaVA-OV and LLaVA-Video.

\end{itemize}

\section{Related Work}

\subsection{Video Large Language Models}

Large vision-language models (LVLMs) combine vision encoders with LLMs for exceptional visual understanding~\cite{li2024llava-ov,Wang:Qwen2-VL}. While LVLMs can handle basic video tasks, the growing demand has led to specialized video large language models (VideoLLMs)~\cite{zhang2024llava-video,Zhang:Video-LLaMA}. These VideoLLMs enhance video understanding through extensive datasets and targeted training strategies, as demonstrated by LLaVA-OneVision~\cite{li2024llava-ov} for multi-modal tasks and LLaVA-Video~\cite{zhang2024llava-video} for video instruction-following. However, the long sequences of visual tokens from continuous video frames limit their practical applications.

\subsection{Token Compression for LVLMs}

Recently, with the increase in visual tokens in LVLMs, research has shifted from \textit{training-aware}~\cite{Li:TokenPacker} to \textit{training-free} token compression methods~\cite{Yang2024:Visionzip}. Training-free approaches are generally categorized as: \textbf{(a)} Pre-LLM token compression at the ViT or projector level~\cite{Zhang:FasterVLM,liu2025globalcom2}; \textbf{(b)} Intra-LLM token compression within the LLM decoder~\cite{Chen:FastV,Zhang:SparseVLM,chen2025v2drop}; and \textbf{(c)} Hybrid token compression that compresses tokens at both ViT and LLM~\cite{Han2024:FiCoCo}. However, these methods treat video frames as separate images, overlooking temporal relationships. While recent work DyCoke~\cite{tao2024dycoke} introduces temporal token merging across consecutive frame windows, it cannot achieve retention ratios below 25\%. More importantly, existing methods, including DyCoke, adopt uniform compression across frames without considering frame uniqueness, and many face compatibility issues with efficient operators~\cite{Dao2022:FlashAttention}.

In this work, we propose a plug-and-play efficient token compression strategy that leverages frame-specific features to tackle current challenges in efficient VideoLLM inference.

\section{Methodology}

\begin{figure*}[t]
    \centering
    \includegraphics[width=\textwidth]{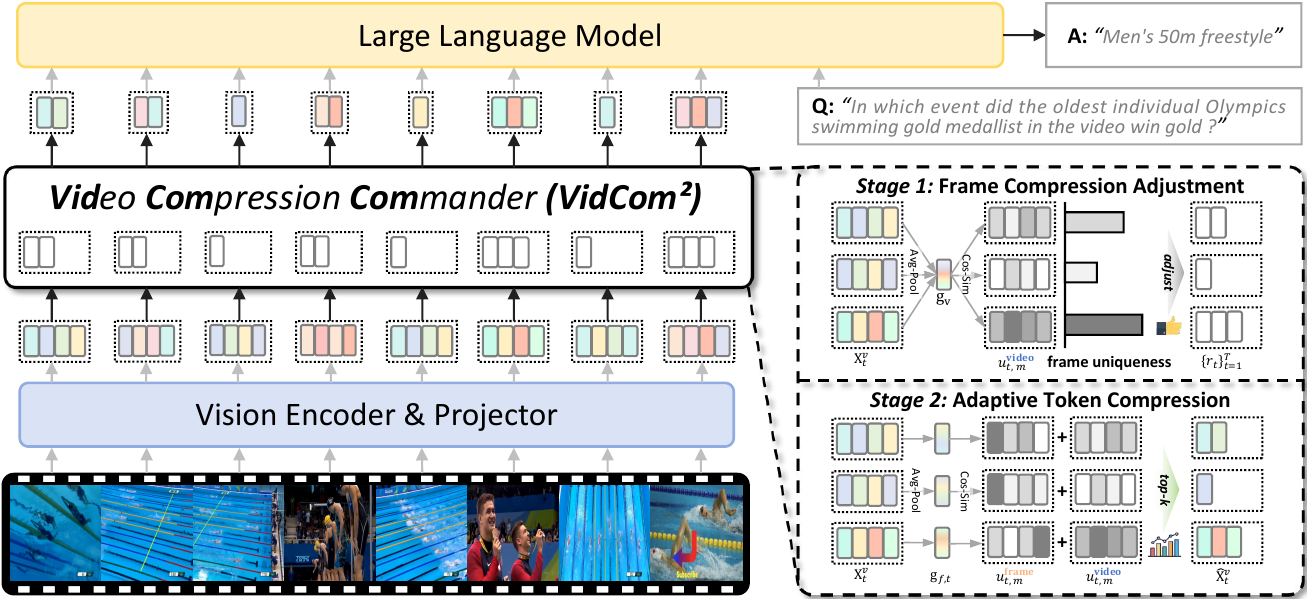}
    \vspace{-7mm}
    \caption{\textbf{Overall framework of VidCom$^2$.} Our VidCom$^2$ performs plug-and-play token compression in two stages: \textbf{(i) Frame Compression Adjustment}: adjusts compression intensity based on frame uniqueness (see Figure~\ref{fig:frame_uniqueness_vidcom2}), \textbf{(ii) Adaptive Token Compression}: preserves tokens based on their within-frame and cross-video uniqueness.}
    \vspace{-5mm}
\label{fig:overview}
\end{figure*}

\subsection{Preliminary}

\paragraph{VideoLLM Architecture.} 

Most current VideoLLMs follow the ``ViT-MLP-LLM'' paradigm~\cite{li2024llava-ov,zhang2024llava-video}. For example, in LLaVA-Video, a video sequence $\mathbf{V} = \left\{ \mathbf{v}_t \right\}_{t=1}^T \in \mathbb{R}^{T \times H \times W \times 3}$ is first encoded by ViT into embeddings $\mathbf{Z} = \left\{ \mathbf{z}_t \right\}_{t=1}^T \in \mathbb{R}^{T \times N \times D}$. These embeddings are projected by a 2-layer MLP and pooled to produce visual tokens $\mathbf{X}^v = \left\{ \mathbf{x}_t^v \right\}_{t=1}^T \in \mathbb{R}^{T \times M \times D'}$, with $M < N$, which are then fed into the LLM for autoregressive instruction-following:
{\setlength\abovedisplayskip{2mm}
\setlength\belowdisplayskip{2mm}
\begin{equation}
p\left(\mathbf{Y} \mid \mathbf{X}^{v}, \mathbf{X}^{t}\right)=\prod_{i=1}^L p\left(\mathbf{y}_i \mid \mathbf{X}^{v}, \mathbf{X}^{t}, \mathbf{Y}_{1:i-1}\right),
\end{equation}
}
where $\mathbf{Y}=\left\{ \mathbf{y}_i \right\}_{i=1}^L$ are the generated response tokens, and $\mathbf{X}^{t}$ are the textual tokens.

\paragraph{Token Compression for VideoLLMs.}

Token compression aims to reduce data redundancy by directly compressing token representations for inference acceleration. For VideoLLMs, this typically involves compressing visual token sequences $\mathbf{X}_{t}^v$ into a reduced representation $\mathbf{\hat{X}}^v$:
\begin{equation}
\mathbf{\hat{X}}^v = \boldsymbol{\Phi}(\mathbf{X}^v), \quad \text{where} \quad |\mathbf{\hat{X}}^v| < |\mathbf{X}^v|
\end{equation}
where $\boldsymbol{\Phi}$ represents the token compression operator and $|\cdot|$ denotes the token length.

Token compression is particularly crucial for VideoLLMs due to their processing of substantially more visual tokens compared to standard LVLMs, a result of the multi-frame nature of videos. Consecutive frames often share high similarity, leading to significant visual redundancy. While recent method DyCoke~\cite{tao2024dycoke} address some aspects of multi-frame redundancy, it struggles with uneven frame distinctiveness and achieving aggressive compression rates. Our work focuses on designing an effective token compression operator $\boldsymbol{\Phi}$ that adaptively handles frame-wise distinctiveness while enabling flexible compression rates, addressing these key challenges for VideoLLMs.

\subsection{Video Compression Commander}

To improve the computational efficiency of VideoLLMs, we propose ``\textbf{Vid}eo \textbf{Com}pression \textbf{Com}mander'' (\textbf{VidCom$^2$}), a novel token compression framework that adaptively minimizes visual redundancy within a predefined token budget while preserving distinctive visual information. VidCom$^2$ maintains compatibility with efficient attention operators~\cite{Dao2022:FlashAttention,daoFlashAttention-2} and supports flexible compression rates, enabling plug-and-play inference acceleration.

Figure~\ref{fig:overview} illustrates the overall framework of VidCom$^2$, which achieves efficient token compression for VideoLLMs through a methodical \textbf{\emph{two-stage}} framework: \textbf{(i) Frame Compression Adjustment}, which evaluates frame uniqueness within the video sequence and dynamically allocates optimal token budgets through compression intensity adjustment; and \textbf{(ii) Adaptive Token Compression}, which assesses token distinctiveness both within-frame and across-video, strategically performing compression based on the frame-specific budgets from the previous stage. Below, we elaborate on the detailed operations of these two stages.

\subsection{Stage 1: Frame Compression Adjustment}

The core of this stage is to adaptively adjust compression intensity based on frame uniqueness across the video. A natural question arises: \textbf{\emph{How can a frame's uniqueness be quantified within the video context?}}. Since each frame $\mathbf{x}_t^v \in \mathbb{R}^{M \times D'}$ consists of $M$ visual tokens, we define frame uniqueness through the collective distinctiveness of its constituent tokens.

\begin{figure*}[!t]
  \centering
   \includegraphics[width=\linewidth]{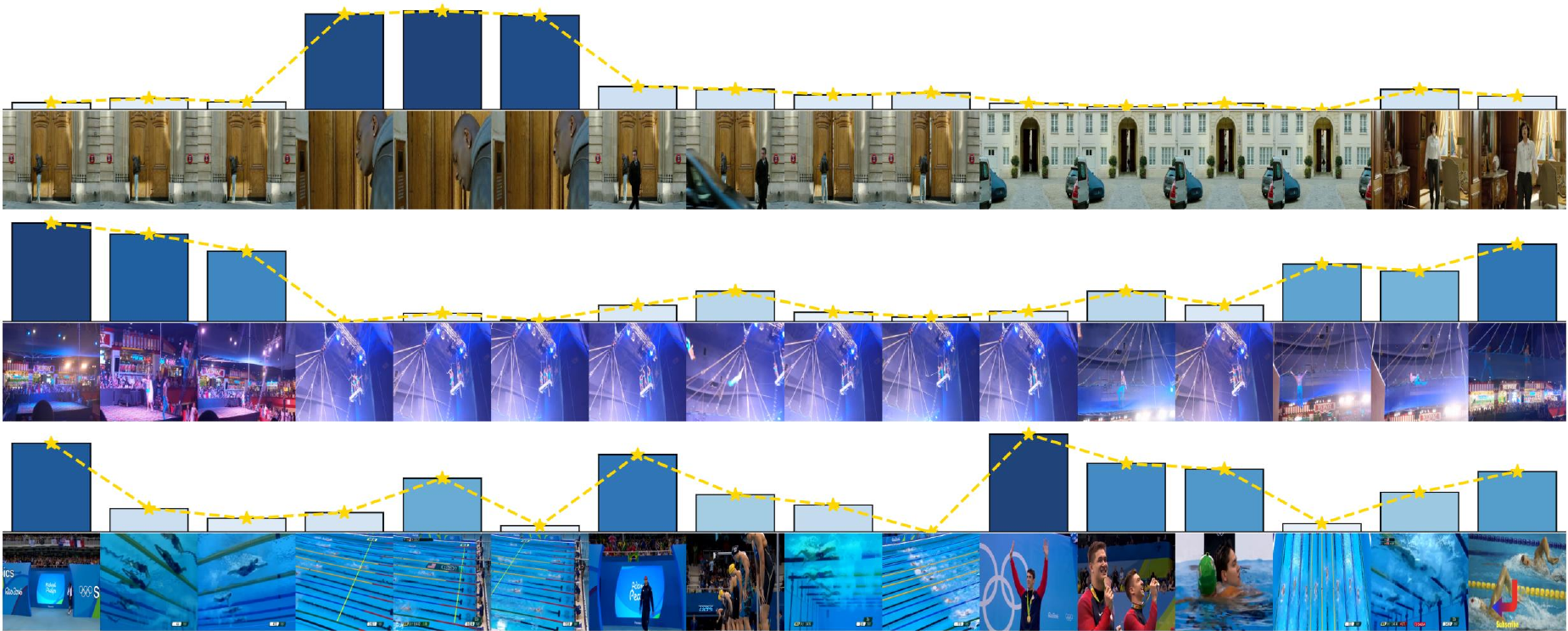}
   \vspace{-7mm}
    \caption{\textbf{Visualization of frame uniqueness quantified by our VidCom$^2$.} Taller and darker bars indicate frame uniqueness, where VidCom$^2$ allocates more tokens to unique frames to preserve critical visual information.}   
   \label{fig:frame_uniqueness_vidcom2}
   \vspace{-5mm}
\end{figure*}

Specifically, we first obtain a global video representation $\mathbf{g_v}$ by average pooling all tokens across $T$ frames, each with $M$ tokens:
{\setlength\abovedisplayskip{2mm}
\setlength\belowdisplayskip{2mm}
\begin{equation}
\mathbf{g_v} = \frac{1}{T \cdot M} \sum_{t=1}^{T} \sum_{m=1}^{M} \mathbf{x}_{t,m}^v,\quad \mathbf{g_v}\in\mathbb{R}^{D'},
\end{equation}}
where $\mathbf{g_v}$ serves as a coarse-grained summary of the entire video. 
Then, inspired by existing efforts~\cite{sun2025mdp3}, we compute the similarity between each token $\mathbf{x}_{t,m}^v$ and global video representation $\mathbf{g_v}$ in high-dimensional space:
{\setlength\abovedisplayskip{2mm}
\setlength\belowdisplayskip{2mm}
\begin{equation}
s^\mathrm{video}_{t,m} = \frac{\mathbf{x}_{t,m}^v \cdot \mathbf{g_v}}{\|\mathbf{x}_{t,m}^v\|\;\|\mathbf{g_v}\|},\quad s^\mathrm{video}_{t,m}\in[-1,1],
\end{equation}}
where a lower $s^\mathrm{video}_{t,m}$ implies that token $\mathbf{x}_{t,m}^v$ is less redundant (more unique) relative to the full video. We define the \emph{video-level uniqueness score} of token $\mathbf{x}_{t,m}^v$ as $u^\mathrm{video}_{t,m} = -s^\mathrm{video}_{t,m}$ and compute the \emph{frame uniqueness score} $u_t = \frac{1}{M}\sum_{m=1}^{M} u^\mathrm{video}_{t,m}$, where a larger $u_t$ indicates higher density of distinctive tokens in frame $t$ compared to the rest of the video. 
Figure~\ref{fig:frame_uniqueness_vidcom2} demonstrates how $u_t$ effectively quantifies frame-wise uniqueness density within video sequences. More cases are in Appendix~\ref{sec:appendix/more_vis}.

These frame-wise scores $\{u_t\}_{t=1}^T$ are used to modulate per-frame compression intensity. To stabilize the scores, we compute $\tilde{u}_t = (u_t - \max(u_t))/\tau$ ($\tau=0.01$), and obtain the relative importance weight $\sigma_t$ of each frame via softmax:
{\setlength\abovedisplayskip{2mm}
\setlength\belowdisplayskip{2mm}
\begin{equation}
    \sigma_t = \frac{\exp(\tilde{u}_t)}{\sum_{l=1}^{T} \exp(\tilde{u}_l) + \epsilon},
\label{eq:frame_relative_importance}
\end{equation}}
where $\epsilon=10^{-8}$ prevents division by zero. Based on these weights, we adjust the preset retention ratio $R (\%)$ for each frame:
{\setlength\abovedisplayskip{2mm}
\setlength\belowdisplayskip{2mm}
\begin{equation}
    r_t = R \times \left(1 + \sigma_t - \frac{1}{T}\right),
\label{eq:frame_ratio_allocation}
\end{equation}}
where $\sigma_t - \frac{1}{T}$ represents the relative deviation from average importance. Consequently, VidCom$^2$ adaptively adjusts compression intensity (\textit{i.e.}, $\{r_t\}_{t=1}^T$) based on frame uniqueness, enabling differentiated token compression degrees across frames while maintaining the average retention ratio $R$.

\subsection{Stage 2: Adaptive Token Compression}

The core of this stage lies in how to select and retain more unique visual information based on the compression degrees $\{r_t\}_{t=1}^T$ determined in the previous stage. Since visual information is composed of tokens, this problem naturally transforms into: \textbf{\emph{How can a token's uniqueness be quantified within the video context?}}. Given the multi-frame nature of videos, a token's uniqueness could be evaluated both locally and globally, \textit{i.e.}, within its frame and across the entire video sequence.

As for token uniqueness within its frame, we can quantify it by measuring its relationship with the frame's global representation. Specifically, for the $t$-th frame, we obtain its global representation through average pooling:
{\setlength\abovedisplayskip{2mm}
\setlength\belowdisplayskip{2mm}
\begin{equation}
\mathbf{g}_{f,t} = \frac{1}{M} \sum_{m=1}^{M} \mathbf{x}_{t,m}^v,\quad \mathbf{g}_{f,t}\in\mathbb{R}^{D'},
\end{equation}}
By computing the cosine similarity of the $m$-th token within the $t$-th frame with its frame-level global representation $\mathbf{g}_{f,t}$, we first define:
{\setlength\abovedisplayskip{2mm}
\setlength\belowdisplayskip{2mm}
\begin{equation}
s^\mathrm{frame}_{t,m} = \frac{\mathbf{x}_{t,m}^v \cdot \mathbf{g}_{f,t}}{\|\mathbf{x}_{t,m}^v\|\;\|\mathbf{g}_{f,t}\|},\quad s_{t,m}\in[-1,1],
\end{equation}}
We then define the \emph{frame-level uniqueness score} as $u^\mathrm{frame}_{t,m} = - s^\mathrm{frame}_{t,m}$, where higher values indicate greater token uniqueness within the frame.

Moreover, since we have already obtained the \emph{video-level uniqueness score} $u^\mathrm{video}_{t,m} = -s^\mathrm{video}_{t,m}$ of token $\mathbf{x}_{t,m}^v$ in the previous stage, we combine these two uniqueness scores to derive \emph{comprehensive uniqueness score} of token $\mathbf{x}_{t,m}^v$ by:
% {\setlength\abovedisplayskip{2mm}
% \setlength\belowdisplayskip{2mm}
% \begin{equation}
% u_{t,m} = \frac{1}{2}\left(u^\mathrm{frame}_{t,m} + u^\mathrm{video}_{t,m}\right),
% \end{equation}}
{\setlength\abovedisplayskip{2mm}
\setlength\belowdisplayskip{2mm}
\begin{equation}
u_{t,m} = u^\mathrm{frame}_{t,m} + u^\mathrm{video}_{t,m},
\end{equation}}
which provides a balanced assessment of the token's distinctiveness both within its frame and across the entire video.

Given the adjusted compression intensity (\textit{i.e.}, $\{r_t\}_{t=1}^T$) based on frame uniqueness in the previous stage, the token compression process for the $t$-th frame can be formulated as:
{\setlength\abovedisplayskip{2mm}
\setlength\belowdisplayskip{2mm}
\begin{equation}
    \mathbf{X}_{t}^{v} \rightarrow \mathbf{\hat{X}}_{t}^{v} = \text{TopK}(\mathbf{X}_{t}^{v}, \{u_{t,m}\}_{m=1}^{M}, r_t \times M)
\end{equation}
}
where $\mathbf{\hat{X}}_{t}^{v}$ represents the compressed token sequence for the $t$-th frame, $\{u_{t,m}\}_{m=1}^{M}$ are the \emph{comprehensive uniqueness scores} of each token in $\mathbf{X}_{t}^{v}$, and $r_t$ is the frame-specific retention ratio. 

To this end, our VidCom$^2$ adaptively adjusts the compression intensity based on frame uniqueness, selectively retaining tokens that are distinctive both within their frames and across the entire video, thereby minimizing information redundancy. The complete algorithm is detailed in Appendix~\ref{sec:appendix/algorithm}.

\section{Experiments}
\label{sec:Experiments}

\begin{table*}[!t]    %[htbp]
\tablestyle{5pt}{1.0}

\newcommand{\downtiny}[1]{{\!\tiny{#1}}}

\setlength\tabcolsep{5pt}
\scalebox{1}{
    \begin{tabular}{lcccccccc}
    \multirow{2}{*}{\textbf{Methods}} & \multirow{2}{*}{\textbf{MVBench}} & \multirow{2}{*}{\textbf{LongVideoBench}} & \multirow{2}{*}{\textbf{MLVU}} & \multicolumn{4}{c}{\textbf{VideoMME}} & \multirow{2}{*}{\textbf{Average (\%)}} \\
        &  &  &  & \textbf{Overall} & \textbf{Short} & \textbf{Medium} & \textbf{Long} &  \\
    \shline
    \multicolumn{9}{l}{\textit{Upper Bound}} \\
    \textcolor[rgb]{ .502,  .502,  .502}{LLaVA-OV-7B} & \textcolor[rgb]{ .502,  .502,  .502}{56.9} & \textcolor[rgb]{ .502,  .502,  .502}{56.4} & \textcolor[rgb]{ .502,  .502,  .502}{63.0} & \textcolor[rgb]{ .502,  .502,  .502}{58.6} & 
    \textcolor[rgb]{ .502,  .502,  .502}{70.3} &
    \textcolor[rgb]{ .502,  .502,  .502}{56.6} &
    \textcolor[rgb]{ .502,  .502,  .502}{48.8} &
    \textcolor[rgb]{ .502,  .502,  .502}{100.0} \\
    \hline
    \multicolumn{9}{l}{\textit{Retention Ratio=30\%}} \\
    DyCoke \downtiny{[CVPR'25]} & 56.6 & 54.7 & 60.3 & 56.1 & 67.1 & 54.6 & 46.6 & 96.5 \\
    \hline
    \multicolumn{9}{l}{\textit{Retention Ratio=25\%}} \\
    Random & 54.2 & 52.7 & 59.7 & 55.6 & 65.4 & 53.0 & 48.3 & 94.8 \\
    \hline
    FastV \downtiny{[ECCV'24]} & 55.5 & 53.3 & 59.6 & 55.3 & 65.0 & 53.8 & 47.0 & 94.9 \\
    PDrop \downtiny{[CVPR'25]} & 55.3 & 51.3 & 57.1 & 55.5 & 64.7 & 53.1 & 48.7 & 94.1 \\
    SparseVLM \downtiny{[ICML'25]} & 56.4 & 53.9 & 60.7 & 57.3 & 68.4 & 55.2  & 48.1 & 97.5 \\
    DyCoke \downtiny{[CVPR'25]} & 49.5 & 48.1 & 55.8 & 51.0 & 61.1 & 48.6 & 43.2 & 87.0 \\
    \rowcolor[rgb]{ .949,  .949,  .949}
    \textbf{VidCom$^2$} & \textbf{57.2} & \textbf{54.9} & \textbf{62.5} & \textbf{58.6} & \textbf{69.8} & \textbf{56.4} & \textbf{49.4} & \textbf{99.6} \\
    \hline
    \multicolumn{9}{l}{\textit{Retention Ratio=15\%}} \\
    FastV \downtiny{[ECCV'24]} & 51.6 & 48.3 & 55.0 & 48.1 & 51.4 & 49.4 & 43.3 & 85.0 \\
    PDrop \downtiny{[CVPR'25]} & 53.2 & 47.6 & 54.7 & 50.1 & 58.7 & 48.7 & 45.0 & 87.4 \\
    SparseVLM \downtiny{[ICML'25]} & 52.9 & 49.7 & 57.4 & 53.4 & 61.0 & 52.1 & 47.0 & 91.2 \\
    \rowcolor[rgb]{ .949,  .949,  .949}
    \textbf{VidCom$^2$} & \textbf{54.3} & \textbf{52.0} & \textbf{58.9} & \textbf{56.2} & \textbf{65.8} & \textbf{54.8} & \textbf{48.1} & \textbf{95.1} \\
    \shline
    \multicolumn{9}{l}{\textit{Upper Bound}} \\
    \textcolor[rgb]{ .502,  .502,  .502}{LLaVA-Video-7B} & \textcolor[rgb]{ .502,  .502,  .502}{60.4} & \textcolor[rgb]{ .502,  .502,  .502}{59.6} & \textcolor[rgb]{ .502,  .502,  .502}{70.3} & \textcolor[rgb]{ .502,  .502,  .502}{64.3} & 
    \textcolor[rgb]{ .502,  .502,  .502}{77.2} &
    \textcolor[rgb]{ .502,  .502,  .502}{62.1} & 
    \textcolor[rgb]{ .502,  .502,  .502}{53.4} & \textcolor[rgb]{ .502,  .502,  .502}{100.0} \\
    \hline
    \multicolumn{9}{l}
    {\textit{Retention Ratio=30\%}} \\
    DyCoke \downtiny{[CVPR'25]} & 57.5 & 55.5 & 60.6 & 61.3 & 73.4 & 59.3 & 51.2 & 93.8 \\
    \hline
    {\textit{Retention Ratio=25\%}} \\
    FastV \downtiny{[ECCV'24]} & 53.8 & 51.2 & 57.8 & 59.3 & 67.1 & 60.0 & \textbf{50.8} & 89.7 \\
    SparseVLM \downtiny{[ICML'25]} & 55.4 & 54.2 & 58.9 & 60.1 & 71.1 & 59.1 & 50.1 & 91.6 \\
    DyCoke \downtiny{[CVPR'25]} & 50.8 & 53.0 & 56.9 & 56.1 & 65.8 & 53.6 & 48.9 & 86.3 \\
    \rowcolor[rgb]{ .949,  .949,  .949}
    \textbf{VidCom$^2$} & \textbf{57.0} & \textbf{55.5} & \textbf{59.0} & \textbf{61.7} & \textbf{73.0} & \textbf{61.7} & 50.0 & \textbf{93.6} \\
    \hline
    \multicolumn{9}{l}{\textit{Retention Ratio=15\%}} \\
    FastV \downtiny{[ECCV'24]} & 44.0 & 44.6 & 53.8 & 51.3 & 56.4 & 51.1 & 46.2 & 78.0 \\
    % PDrop \downtiny{[CVPR'25]} & 49.9 & 48.8 &   & 54.8 & 62.8 & 52.2 & 47.4 &  \\
    SparseVLM \downtiny{[ICML'25]} & 53.1 & \textbf{52.7} & 56.2 & 55.7 & 65.0 & 53.9 & 48.3 & 86.3 \\
    \rowcolor[rgb]{ .949,  .949,  .949}
    \textbf{VidCom$^2$} & \textbf{53.3} & 51.5 & \textbf{56.8} & \textbf{58.3} & \textbf{68.0} & \textbf{57.3} & \textbf{49.7} & \textbf{88.5} \\
    
    \end{tabular}%    
    }
    \vspace{-2mm}
    \caption{\textbf{Performance comparison with other baselines with LLaVA-OV-7B and LLaVA-Video-7B across different benchmarks.} ``Average'' shows the mean performance across different benchmarks. 
    DyCoke requires pruning similar tokens from consecutive 4 frames, making it not possible for the retention ratio of $R<25\%$.
    }
    \vspace{-5mm}
  \label{tab:main_results}%
\end{table*}%

\subsection{Experimental Setting}
\label{subsec:Setting}

\noindent \textbf{Benchmark.} We conduct comprehensive comparative experiments across multiple benchmarks, including: MVBench~\cite{li2024mvbench}, LongVideoBench~\cite{wu2024longvideobench}, MLVU~\cite{zhou2024mlvu}, VideoMME~\cite{fu2024videomme}, EgoSchema~\cite{mangalam2023egoschema}, and PerceptionTest~\cite{patraucean2023perception}, employing LMMs-Eval~\cite{zhang2024lmms} evaluation framework. More details are in Appendix~\ref{sec:appendix/benchmarks}.

\noindent \textbf{Implementations.} We evaluate our method on popular VideoLLMs: LLaVA-OneVision (LLaVA-OV)~\cite{li2024llava-ov}, LLaVA-Video~\cite{zhang2024llava-video}, and Qwen2-VL~\cite{Wang:Qwen2-VL}. Detailed model information is in Appendix~\ref{sec:appendix/models}. All experiments use NVIDIA A100-SXM4-80GB GPUs.

\noindent \textbf{Baselines.} We evaluate our method against various training-free token compression strategies, including: FastV~\cite{Chen:FastV}, PDrop~\cite{xing2024pdrop}, SparseVLM~\cite{Zhang:SparseVLM}, and DyCoke~\cite{tao2024dycoke}, more introduction can be seen in Appendix~\ref{sec:appendix/baselines}. Following SparseVLM, we use the ``equivalent retention ratio''\footnote{``Equivalent retention ratio'' represents the average percentage of visual tokens retained across all LLM layers.} for fair comparisons. Unlike others, DyCoke compresses both visual tokens and KV cache. For fair comparison, we evaluate only on its token compression strategy.

\begin{figure}[t]
    \centering
    \includegraphics[width=\linewidth]{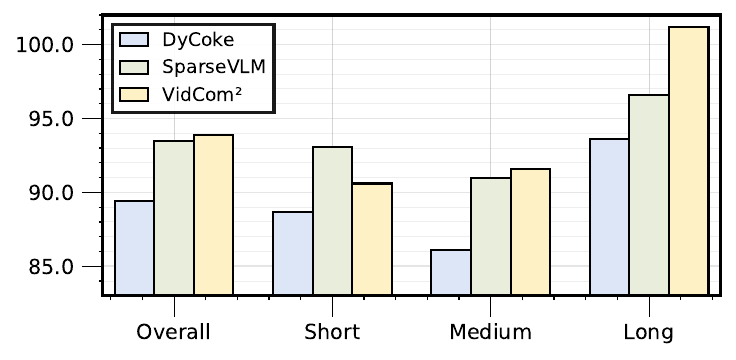}
    \vspace{-7mm}
    \caption{\textbf{Performance with Qwen2-VL.} At $R=25\%$, VidCom$^2$ surpasses DyCoke and SparseVLM by \textbf{7.6\%} and \textbf{4.6\%} of original performance in long video tasks.} 
    \vspace{-5mm}
    \label{fig:qwen_2_vl}
\end{figure}

\subsection{Main Comparisons}
\label{subsec:Comparison}

% \noindent \textbf{Performance Comparisons.}
\paragraph{Performance Comparisons.} 

Table~\ref{tab:main_results} presents a comparative analysis of our VidCom$^2$ against multiple token compression methods across various benchmarks. The experimental results reveal \textbf{\emph{two key performance advantages}} of VidCom$^2$:

\begin{table}[!t]
\tablestyle{5pt}{1.0}
\newcommand{\downtiny}[1]{{\!\tiny{#1}}}
\setlength\tabcolsep{8.8pt}
\scalebox{1}{
    \begin{tabular}{lcc}
    \textbf{Methods} & \textbf{EgoSchema} & \textbf{PerceptionTest} \\
    \shline
    \multicolumn{3}{l}{\textit{Upper Bound}} \\
    \textcolor[rgb]{.502,.502,.502}{LLaVA-OV-7B} & \textcolor[rgb]{.502,.502,.502}{60.4 (100\%)} & \textcolor[rgb]{.502,.502,.502}{57.1 (100\%)} \\
    \hline
    \multicolumn{3}{l}{\textit{Retention Ratio=25\%}} \\
    FastV \downtiny{[ECCV'24]} & 57.5 (95.2\%) & 55.4 (97.0\%) \\
    PDrop \downtiny{[CVPR'25]} & 58.0 (96.0\%) & 55.6 (97.4\%) \\
    DyCoke \downtiny{[CVPR'25]} & 59.5 (98.5\%) & 56.4 (98.8\%) \\
    \rowcolor[rgb]{.949,.949,.949}
    \textbf{VidCom$^2$} & \textbf{59.7 (98.8\%)} & \textbf{56.7 (99.3\%)} \\
    \end{tabular}
}
\vspace{-2mm}
\caption{\textbf{Performance comparison on EgoSchema and PerceptionTest.} Percentages represent ratios to the original performance of LLaVA-OV-7B.}
\vspace{-5mm}
\label{tab:more_comparisons}
\end{table}

\definecolor{greenrightcolor}{RGB}{0,144,81} 
\definecolor{redrightcolor}{RGB}{255,0,0} 
\definecolor{bluerightcolor}{RGB}{0,0,255} % 

\newcommand{\downtinya}[1]{{\!\textcolor{greenrightcolor}{\tiny{#1}}}}
\newcommand{\uptiny}[1]{{\!\textcolor{redrightcolor}{\tiny{#1}}}}
\newcommand{\downtinyb}[1]{{\!\textcolor{bluerightcolor}{\tiny{#1}}}}

\newcommand{\downtiny}[1]{{\!\tiny{#1}}}

\begin{table*}[!t]
\tablestyle{5pt}{1.0}
  \centering
  \setlength{\tabcolsep}{0.5pt} 
    \scalebox{1}{
    \begin{tabular}{lccccccc}
    \multirow{2}{*}{\textbf{Methods}} & \textbf{LLM Generation↓} & \textbf{Model Generation↓} & \textbf{Total↓}  & \textbf{GPU Peak↓} & \textbf{Throughput↑} & \multirow{2}{*}{\textbf{Performance↑}} \\
    & \textbf{Latency (s)} & \textbf{Latency (s)} & \textbf{Latency (min:sec)} & \textbf{Memory (GB)} & \textbf{(samples/s)} &  \\
    \shline
    \textcolor[rgb]{ .502,  .502,  .502}{LLaVA-OV-7B} & \textcolor[rgb]{ .502,  .502,  .502}{618.0} & \textcolor[rgb]{ .502,  .502,  .502}{1008.4} & 
    \textcolor[rgb]{ .502,  .502,  .502}{26:03} & 
    \textcolor[rgb]{ .502,  .502,  .502}{17.7} & 
    \textcolor[rgb]{ .502,  .502,  .502}{0.64} & \textcolor[rgb]{ .502,  .502,  .502}{56.9} \\
    
    \hline
    \multicolumn{7}{l}{\textit{Retention Ratio=25\%}} \\
    Random  & 178.2 \downtinya{(↓71.2\%)} & 566.0 \downtinya{(↓43.9\%)} & 18:44 \downtinya{(↓28.1\%)} & 16.0 \downtinya{(↓9.6\%)} & 0.89 \downtinya{(1.39$\times$)} & 54.6 \downtinya{(↓2.3)} \\

    \hline
    
    FastV \downtiny{[ECCV'24]}  & 260.9 \downtinya{(↓57.8\%)} & 648.6 \downtinya{(↓35.7\%)} & 20:07 \downtinya{(↓22.8\%)} & 24.7 \uptiny{(↑39.5\%)} & 0.83 \downtinya{(1.30$\times$)} &  55.5 \downtinya{(↓1.4)}  \\
    
    PDrop \downtiny{[CVPR'25]}  & 205.6 \downtinya{(↓66.7\%)} & 592.6 \downtinya{(↓41.2\%)} & 18:50 \downtinya{(↓27.7\%)} & 24.5 \uptiny{(↑38.4\%)} & 0.88 \downtinya{(1.38$\times$)} &  55.3 \downtinya{(↓1.6)}  \\
    
    SparseVLM \downtiny{[ICML'25]}  & 410.6 \downtinya{(↓33.6\%)} & 807.7 \downtinya{(↓19.9\%)} & 25:03 \downtinya{(↓3.8\%)} & 27.1 \uptiny{(↑53.1\%)} & 0.67 \downtinya{(1.05$\times$)} & 56.4 \downtinya{(↓0.5)} \\
    
    DyCoke \downtiny{[CVPR'25]}  & 205.2 \downtinya{(↓66.8\%)} & 598.0 \downtinya{(↓40.7\%)} & 18:56 \downtinya{(↓27.4\%)} & 16.1 \downtinya{(↓9.0\%)} & 0.88 \downtinya{(1.38$\times$)} & 49.5 \downtinya{(↓7.4)} \\
    
    \rowcolor[rgb]{ .949,  .949,  .949} \textbf{VidCom$^2$}  & \textbf{180.7 \downtinya{(↓70.8\%)}} & \textbf{574.7 \downtinya{(↓43.0\%)}} & \textbf{18:46 \downtinya{(↓28.0\%)}} & \textbf{16.0 \downtinya{(↓9.6\%)}} & \textbf{0.88 \downtinya{(1.38$\times$)}} & \textbf{57.2 \uptiny{(↑0.3)}} \\
    
    \hline
    \multicolumn{7}{l}{\textit{Retention Ratio=15\%}} \\
    Random  & 130.3 \downtinya{(↓78.9\%)} & 532.5 \downtinya{(↓47.2\%)} & 18:02 \downtinya{(↓30.8\%)} & 15.8 \downtinya{(↓10.7\%)} & 0.92 \downtinya{(1.44$\times$)} & 53.1 \downtinya{(↓3.8)} \\

    \hline
    
    FastV \downtiny{[ECCV'24]}  & 172.4 \downtinya{(↓72.1\%)} & 599.3 \downtinya{(↓40.6\%)} & 18:19 \downtinya{(↓29.7\%)} & 24.6 \uptiny{(↑39.0\%)} & 0.91 \downtinya{(1.42$\times$)} & 51.6 \downtinya{(↓5.3)} \\
    
    PDrop \downtiny{[CVPR'25]}  & 165.3 \downtinya{(↓73.3\%)} & 552.6 \downtinya{(↓45.2\%)} & 18:32 \downtinya{(↓28.9\%)} & 24.5 \uptiny{(↑38.4\%)} & 0.90 \downtinya{(1.41$\times$)} & 53.2 \downtinya{(↓3.7)} \\

    SparseVLM \downtiny{[ICML'25]}  & 370.4 \downtinya{(↓40.1\%)} & 764.8 \downtinya{(↓24.2\%)} & 24:09 \downtinya{(↓7.3\%)} & 27.1 \uptiny{(↑53.1\%)} & 0.69 \downtinya{(1.08$\times$)} & 52.9 \downtinya{(↓4.0)} \\
    
    \rowcolor[rgb]{ .949,  .949,  .949} \textbf{VidCom$^2$}  & \textbf{129.2 \downtinya{(↓79.1\%)}} & \textbf{533.0 \downtinya{(↓47.1\%)}} & \textbf{18:11 \downtinya{(↓30.2\%)}} & \textbf{15.8 \downtinya{(↓10.7\%)}} & \textbf{0.92 \downtinya{(1.44$\times$)}} & \textbf{54.3 \downtinya{(↓2.6)}} \\
    
    \end{tabular}%
    }
    \vspace{-2mm}
  \caption{\textbf{Efficiency comparisons on LLaVA-OV-7B.} ``LLM Generation Latency'': time for LLM-only response generation; ``Model Generation Latency'': time for model to generate response; ``Total Latency'': total time to complete MVBench; and  ``Throughput'': number of MVBench samples processed per second.}
  \label{tab:efficiency_comparisons}%
  \vspace{-5mm}
\end{table*}%

\noindent \textbf{(i) State-of-the-art Performance:} VidCom$^2$ demonstrates exceptional performance across diverse video understanding benchmarks. On LLaVA-OV and LLaVA-Video with compression ratio $R=25\%$, VidCom$^2$ substantially outperforms DyCoke by margins of \textbf{12.6\%} and \textbf{7.3\%}, respectively. Remarkably, VidCom$^2$ at $R=25\%$ (achieving 99.6\% performance retention) even surpasses DyCoke operating at a higher compression ratio of $R=30\%$ (96.5\% performance retention). This superiority extends to long-form video understanding tasks with Qwen2-VL (Figure~\ref{fig:qwen_2_vl}), where VidCom$^2$ achieves \textbf{101.2\%} performance on VideoMME (Long), surpassing both DyCoke (93.6\%) and SparseVLM (96.6\%) by substantial margins of \textbf{7.6\%} and \textbf{4.6\%}, respectively. Additional comparisons in Table~\ref{tab:more_comparisons} further validate the superior performance advantages of VidCom$^2$ across various video understanding scenarios.

\noindent \textbf{(ii) Robustness in Extreme Compression:} Under aggressive compression with $R=15\%$, most baselines such as FastV and PDrop exhibit significant performance degradation. Even the VideoLLM-specific method DyCoke \textbf{fails to achieve} such aggressive compression due to inherent design limitations.
However, VidCom$^2$ maintains robust performance, outperforming the second-best method SparseVLM by an average of \textbf{3.9\%} and \textbf{2.1\%} on LLaVA-OV and LLaVA-Video. 
This demonstrates VidCom$^2$'s superiority in frame-adaptive compression, dynamically adjusting intensity to preserve distinctive visual information.

Besides, we observe an interesting phenomenon that Intra-LLM methods (\textit{e.g.}, SparseVLM), which incorporate textual information, perform relatively better on long video tasks (\textit{e.g.}, LongVideoBench and VideoMME (long)) compared to shorter video benchmarks like MVBench and VideoMME (Short). For instance, SparseVLM slightly outperforms VidCom$^2$ on LongVideoBench with LLaVA-Video at $R=15\%$. This suggests that for longer videos with fixed frame counts, leveraging textual information for visual token compression helps VideoLLMs focus on text-relevant visual areas, potentially leading to improved performance.

% \noindent \textbf{Efficiency Comparisons.} 
\paragraph{Efficiency Comparisons.}

Beyond performance, Table~\ref{tab:efficiency_comparisons} presents comprehensive real-world inference efficiency comparisons among different token compression methods on MVBench, with all experiments conducted on four NVIDIA A100 GPUs. We follow the original implementation of each baseline method, and unless otherwise specified, Flash Attention 2~\cite{daoFlashAttention-2} is used as the efficient attention operator throughout comparisons. The comparison results in Table~\ref{tab:efficiency_comparisons} reveal \textbf{\emph{two key efficiency advantages}} of our VidCom$^2$:

\noindent \textbf{(i) State-of-the-art Efficiency:} VidCom$^2$ achieves remarkable inference efficiency, \textbf{comparable to} simple random token dropping. With 25\% visual tokens retained, the additional computation of VidCom$^2$ is \textbf{negligible} -- only 2.5s extra (\textbf{1.3\%} of LLM generation time) for the entire MVBench inference. Despite this minimal overhead, VidCom$^2$ significantly reduces both the LLM generation latency and overall model latency (primarily from ViT and LLM) by \textbf{70.8\%} and \textbf{43.0\%} respectively, achieving \textbf{1.38$\times$} throughput while maintaining \textbf{99.6\%} average performance across benchmarks. These results highlight the efficiency of VidCom$^2$ in accelerating inference for VideoLLMs.

\noindent \textbf{(ii) Efficient Operator Compatibility:} Pre-LLM methods like DyCoke and our VidCom$^2$ maintain Flash Attention compatibility while continuously reducing peak memory usage, showcasing their efficiency. When equipped with Flash Attention, both VidCom$^2$ and random dropping further reduce peak memory usage by approximately 2 GB compared to standard Flash Attention, demonstrating that VidCom$^2$'s computation introduces no additional memory overhead. In contrast, Intra-LLM methods (\textit{e.g.}, PDrop and FastV) even substantially \textbf{increase} memory consumption. For instance, FastV increases the original peak memory by significantly \textbf{39.5\%}. This dramatic increase stems from their reliance on \textbf{explicit} attention weights, rendering them incompatible with Flash Attention in certain layers. Given the large number of frames and tokens in video sequences, such memory-intensive methods show limited practical value for VideoLLMs.

\begin{table}[!t]
  \centering
  \setlength{\tabcolsep}{1.5pt}
  \scalebox{0.8}{
    \begin{tabular}{lcccccc}
    \multirow{2}{*}{\textbf{Metrics}} & \multirow{2}{*}{\textbf{MLVU}}  &  \multicolumn{4}{c}{\textbf{VideoMME}} & \multirow{2}{*}{\textbf{Avg.}} \\
        & & \textbf{Overall} & \textbf{Short} & \textbf{Medium} & \textbf{Long}  & \\
    \shline
    \textcolor[rgb]{ .502,  .502,  .502}{Vanilla} & 
    \textcolor[rgb]{ .502,  .502,  .502}{63.0} &
    \textcolor[rgb]{ .502,  .502,  .502}{58.6} & 
    \textcolor[rgb]{ .502,  .502,  .502}{70.3} &
    \textcolor[rgb]{ .502,  .502,  .502}{56.6} &
    \textcolor[rgb]{ .502,  .502,  .502}{48.8} &
    \textcolor[rgb]{ .502,  .502,  .502}{100.0}
    \\
    \hline
    $s^\mathrm{frame}_{t,m}$ & 59.5  & 54.0 & 62.2 &  54.2 & 45.3 & 94.1  \\
    \rowcolor[rgb]{ .949,  .949,  .949}
    $- s^\mathrm{frame}_{t,m}$ & 61.9  & 57.9  & 68.8  & 56.9  & 48.1 & 98.8 \\
    \hline
    $s^\mathrm{video}_{t,m}$ &  58.9 & 53.3  & 61.7  &  52.1 & 46.1 & 93.2 \\
    \rowcolor[rgb]{ .949,  .949,  .949}
    $-s^\mathrm{video}_{t,m}$  & 61.4  &  58.3 & 69.3  &  56.1 & 49.3 & 99.3 \\
    \hline
    \rowcolor[rgb]{ .949,  .949,  .949}
    \textbf{$u^\mathrm{frame}_{t,m} + u^\mathrm{video}_{t,m}$}  & \textbf{62.1} & \textbf{58.5} & \textbf{69.6} & \textbf{56.3} & \textbf{49.3} & \textbf{99.7} \\
    \end{tabular}%
    }
  \vspace{-2mm}
  \caption{\textbf{Effects of different token evaluation metrics.} The first two parts explores the optimal $u^\mathrm{frame}_{t,m}$ and $u^\mathrm{video}_{t,m}$, while the last part examines the optimal $u_{t,m}$.}
  \label{tab:ablation_2}%
  \vspace{-5mm}
\end{table}%

\subsection{Ablation Study and Analysis}
\label{subsec:Ablation}

We conduct multiple ablation studies and analyses with $R = 25\%$ on LLaVA-OV-7B, exploring optimal token evaluation strategies and validating the effectiveness of Frame Compression Adjustment for both VidCom$^2$ and other methods.

% \noindent \textbf{Ablation on Different Token Evaluation.} 

\paragraph{Effects of Different Token Evaluation Metrics.}

Table~\ref{tab:ablation_2} presents various metrics for token evaluation, consisting of three parts: \textbf{(a)} frame-level uniqueness score $u^\mathrm{frame}_{t,m}$, \textbf{(b)} video-level uniqueness score $u^\mathrm{video}_{t,m}$, and \textbf{(c)} the final score $u_{t,m}$ that combines $u^\mathrm{frame}_{t,m}$ and $u^\mathrm{video}_{t,m}$ to guide our token preservation strategy.

For frame-level uniqueness, defining $u^\mathrm{frame}_{t,m}$ as the negative similarity to frame-level global representation ($-s^\mathrm{frame}_{t,m}$) outperforms positive similarity. Similarly, for video-level uniqueness, tokens less similar to the video-level global representation prove more informative. These results indicate that unique tokens, both within frames and across the video, should be prioritized during token compression to preserve richer visual information.

Token compression guided by either frame-level or video-level uniqueness scores outperforms the baselines in Table~\ref{tab:main_results}, showcasing the effectiveness of uniqueness-based selection. Their combination further achieves optimal performance, suggesting that token uniqueness should be evaluated both \textbf{within-frame} and \textbf{across-video} to maximize visual content preservation during token compression.

\begin{table}[!t]
  \centering
  \setlength{\tabcolsep}{3pt}
  \scalebox{0.8}{
    \begin{tabular}{lcccccc}
    \multirow{2}{*}{\textbf{Metrics}} &  \multirow{2}{*}{\textbf{MLVU}}  &  \multicolumn{4}{c}{\textbf{VideoMME}} & \multirow{2}{*}{\textbf{Avg.}} \\
        & & \textbf{Overall} & \textbf{Short} & \textbf{Medium} & \textbf{Long}  & \\
    \shline
    \textcolor[rgb]{ .502,  .502,  .502}{Vanilla} & 
    \textcolor[rgb]{ .502,  .502,  .502}{63.0} &
    \textcolor[rgb]{ .502,  .502,  .502}{58.6} & 
    \textcolor[rgb]{ .502,  .502,  .502}{70.3} &
    \textcolor[rgb]{ .502,  .502,  .502}{56.6} &
    \textcolor[rgb]{ .502,  .502,  .502}{48.8} &
    \textcolor[rgb]{ .502,  .502,  .502}{100.0}
    \\
    \hline
    Uniform  & 61.9  &  57.9 & 68.8  & \textbf{56.9}  & 48.1 & 98.8  \\
    \hline
    \multicolumn{7}{l}{\textit{Frame Compression Adjustment}} \\
    $\max{u^\mathrm{video}_{t,m}}$ & 62.1  & 58.1  &  68.4 & 56.7  & 49.3 & 99.4 \\
    \rowcolor[rgb]{ .949,  .949,  .949}
    $\overline{u^\mathrm{video}_{t,m}}$  & \textbf{62.3}  & \textbf{58.2} & \textbf{69.1}  & 55.9  & \textbf{49.6} & \textbf{99.6} \\
    \end{tabular}%
    }
  \vspace{-2mm}
  \caption{\textbf{Effects of different compression adjustment.} ``Uniform'': fixed $R=25\%$. ``$\max u^\mathrm{video}_{t,m}$'' and ``$\overline{u^\mathrm{video}_{t,m}}$'' denote frame uniqueness score $u_t$ of frame $t$ computed by maximum and average operations of $u^\mathrm{video}_{t,m}$.}
  \label{tab:ablation_1_ours}%
  \vspace{-3mm}
\end{table}%

% \noindent \textbf{Ablation on Frame Compression Adjustment.} 
\paragraph{Effects of Frame Compression Adjustment.} Table~\ref{tab:ablation_1_ours} compares different compression adjustment strategies: \textbf{(a)} ``Uniform'' with fixed $R=25\%$ (no adjustment); \textbf{(b)} ``$\max\limits_{m} u^\mathrm{video}_{t,m}$'' and \textbf{(c)} ``$\overline{u^\mathrm{video}_{t,m}}$'', which compute frame uniqueness score $u_t$ for token budget allocation using maximum and average operations of $u^\mathrm{video}_{t,m}$ in frame $t$, where larger $u_t$ leads to more tokens preserved in frame $t$.

Generally, Frame Compression Adjustment strategies demonstrate performance improvements over uniform compression, validating the effectiveness of dynamically adjusting compression intensity based on frame uniqueness. This confirms our intuition that allocating more token budget to distinctive frames helps preserve important visual information along the temporal dimension. Moreover, averaging token uniqueness ($\overline{u^\mathrm{video}_{t,m}}$) outperforms maximum operation ($\max\limits_{m} u^\mathrm{video}_{t,m}$), as it better captures the overall \textbf{uniqueness density} of a frame rather than focusing on isolated distinctive features, providing a more comprehensive measure of frame-level temporal uniqueness.

\begin{table}[!t]
  \centering
  \setlength{\tabcolsep}{3pt}
  \scalebox{0.8}{
    \begin{tabular}{lcccccc}
    \multirow{2}{*}{\textbf{Size}} &  \multirow{2}{*}{\textbf{MVBench}}  &  \multicolumn{4}{c}{\textbf{VideoMME}} & \multirow{2}{*}{\textbf{Avg.}} \\
        & & \textbf{Overall} & \textbf{Short} & \textbf{Medium} & \textbf{Long}  & \\
    \shline
    \textcolor[rgb]{ .502,  .502,  .502}{Vanilla} & 
    \textcolor[rgb]{ .502,  .502,  .502}{56.9} &
    \textcolor[rgb]{ .502,  .502,  .502}{58.6} & 
    \textcolor[rgb]{ .502,  .502,  .502}{70.3} &
    \textcolor[rgb]{ .502,  .502,  .502}{56.6} &
    \textcolor[rgb]{ .502,  .502,  .502}{48.8} &
    \textcolor[rgb]{ .502,  .502,  .502}{100.0} \\
   \hline
    4 & 56.8 & 57.9 & 69.6 & 55.6 & 48.7 & 99.1 \\
    8 & 56.8 & 58.3 & 69.8 & 56.4 & 48.6 & 99.6 \\
    16 & 57.2 & 58.5 & \textbf{70.0} & \textbf{56.7} & 48.9 & 100.1 \\
    \rowcolor[rgb]{ .949,  .949,  .949}
    32 & \textbf{57.2} & \textbf{58.6} & 69.8 & 56.4 & \textbf{49.4} & \textbf{100.1} \\
    \end{tabular}%
    }
  \vspace{-2mm}
  \caption{\textbf{Effects of different window sizes for local $g_v$ computation.} Window sizes up to 32 (global perspective) are evaluated on LLaVA-OV-7B.}
  \label{tab:ablation_window_size}%
  \vspace{-5mm}
\end{table}%

\paragraph{Effects of Different Window Sizes for Local $g_v$ Computation}
We explore sliding window strategies for computing local $g_v$ representations to investigate the effectiveness of adjusting frame compression intensity from local perspectives. We evaluate different window sizes (4, 8, 16, 32) on LLaVA-OV-7B with fixed 32 frames.

As shown in Table~\ref{tab:ablation_window_size}, performance consistently improves as window size increases across both MVBench and VideoMME. Notably, when window sizes reach 16 and 32, the performance gap becomes marginal. Window size 16 achieves better results on VideoMME short and medium videos, while window size 32 (global perspective) demonstrates superior performance on VideoMME long videos. Therefore, we adopt the global perspective for adjusting compression intensity by default to achieve better long video understanding.

\begin{figure}[t]
    \centering
    \includegraphics[width=\linewidth]{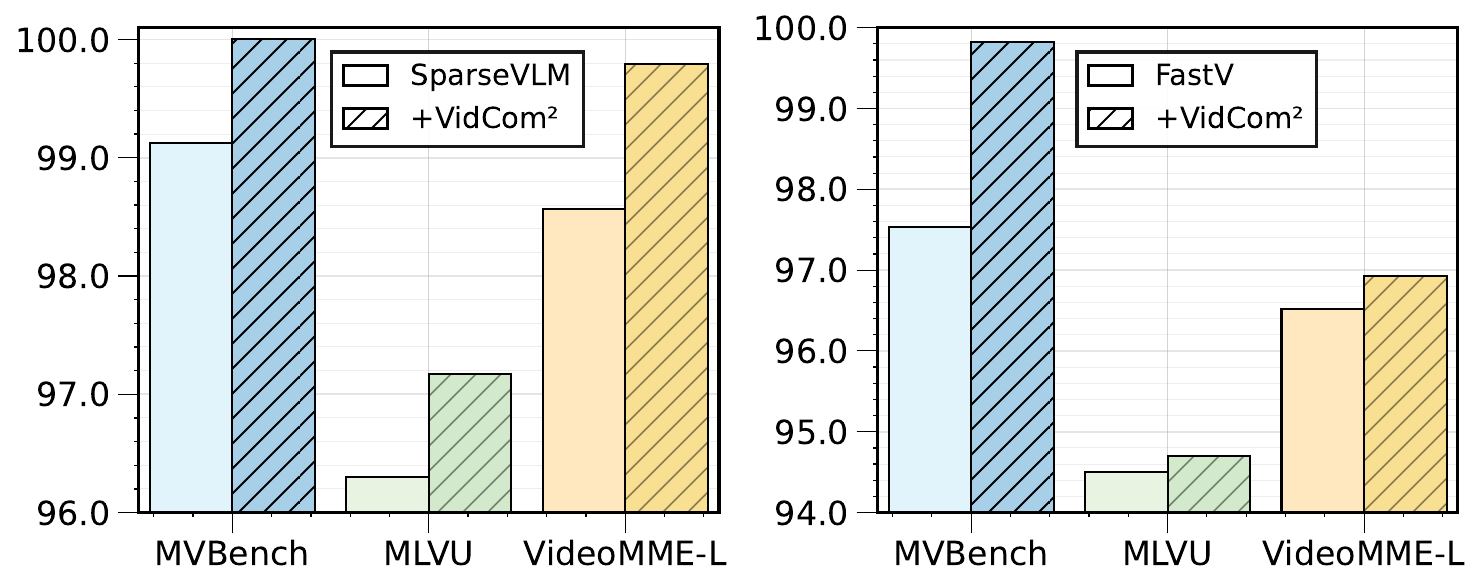}
    \vspace{-7mm}
    \caption{\textbf{Effects of Frame Compression Adjustment on other methods.}``+VidCom$^2$'' indicates the application of our Frame Compression Adjustment strategy.} 
    \vspace{-5mm}
\label{fig:ablation_1_others}
\end{figure}

% \noindent \textbf{Broader Applicability of VidCom$^2$.} 
\paragraph{Broad Applicability of Frame Compression Adjustment.}

Figure~\ref{fig:ablation_1_others} further demonstrates the effectiveness of integrating our Frame Compression Adjustment strategy with other methods. 

Results show consistent performance improvements compared to their original performance on LLaVA-OV-7B across short (MVBench) and long (MVLU and VideoMME-L) video understanding tasks. Notably, SparseVLM and FastV show significant gains on MVBench, where spatiotemporal changes are more pronounced. This improvement stems from the complementary nature of our approach: while Intra-LLM methods focus on \textbf{textual relevance}, our strategy considers \textbf{visual uniqueness}. This combination enables more comprehensive token preservation, capturing both distinctive visual content and instruction-relevant information, thus mitigating unique visual information loss that often occurs in text-centric approaches during token compression in LLM.

\section{Conclusion}
\label{sec:Conclusion}

In this work, we first analyze existing token compression methods for VideoLLMs, identifying two key limitations: design myopia and implementation constraints. We then derive three principles for effective token compression: model adaptability, frame uniqueness, and operator compatibility. Guided by the three principles, we propose VidCom$^2$, a novel plug-and-play acceleration framework. VidCom$^2$ dynamically adjusts compression intensity based on frame uniqueness, effectively preserving the most distinctive tokens both within each frame and across the entire video. Extensive experiments demonstrate VidCom$^2$ achieves state-of-the-art performance and efficiency across diverse benchmarks.

\section{Limitations}
\label{sec:Limitations}

In our work, we propose a plug-and-play efficient token compression framework for VideoLLM acceleration. Due to computational constraints, we couldn't evaluate our method on larger models like LLaVA-Video-72B and Qwen2-VL-72B. However, given VidCom$^2$'s simplicity and the significant advantages demonstrated in Table~\ref{tab:main_results} and Table~\ref{tab:efficiency_comparisons}, we anticipate its benefits may extend to or even amplify in larger architectures. This expectation is based on the increased importance of efficient token management in more complex models. Future work will focus on comprehensive evaluation across various model sizes to further validate and explore VidCom$^2$'s potential in larger-scale scenarios. Additionally, we aim to adapt VidCom$^2$ for real-time streaming video understanding scenarios, further expanding its practical applications.

\section*{Acknowledgement}

This research was supported by the Shanghai Science and Technology Program (Grant No. 25ZR1402278). Besides, we thank Huawei Ascend Cloud Ecological Development Project for the support of Ascend 910 processors.

\bibliography{custom}

\clearpage
\appendix

% \section{Discussion of Similarity Computation}
% \label{sec:appendix/similarity}

In the appendix, we provide more benchmark details in Section~\ref{sec:appendix/benchmarks}, model details in Section~\ref{sec:appendix/models}, more baseline details in Section~\ref{sec:appendix/baselines}, sensitivity analysis in Section~\ref{sec:appendix/hyper-para}, algorithm details in Section~\ref{sec:appendix/algorithm}, and more visualization of frame uniqueness quantified by VidCom$^2$ in Section~\ref{sec:appendix/more_vis}.

\section{Benchmark Details}
\label{sec:appendix/benchmarks}

We present a detailed overview of video understanding benchmarks, as described below:

We present a detailed overview of video understanding benchmarks, as described below:
\begin{itemize}
    \item \textbf{MVBench}~\cite{li2024mvbench} defines 20 video understanding tasks that require deep comprehension of temporal dimensions, beyond single-frame analysis.
    
    \item \textbf{LongVideoBench}~\cite{wu2024longvideobench} focuses on long-context video understanding with 3,763 videos up to one hour long. It includes 6,678 multiple-choice questions across 17 categories, emphasizing temporal information retrieval and analysis.
    
    \item \textbf{MLVU}~\cite{zhou2024mlvu} features videos ranging from 3 minutes to 2 hours, encompassing 9 evaluation tasks including topic reasoning, anomaly recognition, video summarization, and plot question-answering.
    
    \item \textbf{VideoMME}~\cite{fu2024videomme} comprises 900 videos and 2,700 multiple-choice questions across six domains, with durations from 11 seconds to 1 hour, categorized into short, medium, and long subsets.
    
    \item \textbf{EgoSchema}~\cite{mangalam2023egoschema} consists of 5,000 egocentric videos with multiple-choice questions requiring comprehensive understanding of procedural activities and temporal reasoning over extended sequences, challenging models with first-person perspective video analysis.
    
    \item \textbf{PerceptionTest}~\cite{patraucean2023perception} presents 11,609 real-world videos with 38,565 multiple-choice questions evaluating diverse perceptual skills including object tracking, action recognition, and temporal localization across varied scenarios and contexts.
    
\end{itemize}

\section{Model Details}
\label{sec:appendix/models}

We introduce the VideoLLMs used for evaluation in main text, as follows:

\begin{itemize}

    \item \textbf{LLaVA-OneVision}~\cite{li2024llava-ov} unifies single-image, multi-image, and video tasks in a single LLaVA-OneVision model. It represents videos as long visual token sequences in the same ``interleaved'' format used for images, enabling smooth task transfer from images to videos and facilitating strong zero-shot video understanding capabilities.
    
    \item \textbf{LLaVA-Video}~\cite{zhang2024llava-video} builds upon the single-image stage checkpoint of LLaVA-OneVision. It is fine-tuned on a large synthetic video-instruction dataset (LLaVA-Video-178K), covering detailed captioning, open-ended QA, and multiple-choice QA. By employing the SigLIP visual encoder and Qwen2 as the LLM, LLaVA-Video achieves robust video comprehension across various benchmarks.
    
    \item \textbf{Qwen2-VL}~\cite{Wang:Qwen2-VL} introduces Naive Dynamic Resolution to adaptively convert frames of any resolution into visual tokens. It utilizes Multimodal Rotary Position Embedding within a unified image-and-video processing paradigm, enabling the handling of long videos (20+ minutes) for high-quality QA, dialogue, and content creation.

\end{itemize}

\section{Baseline Details}
\label{sec:appendix/baselines}

We provide detailed introductions and comparisons of existing token compression methods mentioned in the main text, as follows:

\begin{itemize}

    \item \textbf{FastV}~\cite{Chen:FastV} performs one-time token pruning as an intra-LLM compression method, utilizing attention weights associated with the output token after a selected LLM layer. However, its explicit dependence on attention weights makes it incompatible with Flash Attention~\cite{Dao2022:FlashAttention} in LLM.

    \item \textbf{PDrop}~\cite{xing2024pdrop} extends intra-LLM compression by implementing progressive token pruning across multiple LLM layers, based on attention weights of output tokens. Similarly, this explicit attention mechanism prevents compatibility with Flash Attention~\cite{Dao2022:FlashAttention} in LLM.
    
    \item \textbf{SparseVLM}~\cite{Zhang:SparseVLM} functions as an intra-LLM compression method, ranking token importance using text-visual attention maps and pruning via pre-selected text prompts to mitigate attention noise. Similar to FastV, SparseVLM is also incompatible with Flash Attention~\cite{Dao2022:FlashAttention} in LLM.

    \item \textbf{MUSTDrop}~\cite{Liu2024:MUSTDrop} is a three-stage compression method operating in ViT and LLM stages. It relies on \texttt{[CLS]} token attention and text-visual attention for token selection and pruning. This approach faces compatibility issues with \texttt{[CLS]}-free VideoLLMs and prevents Flash Attention support in LLM due to its explicit use of attention weights.
    
    \item \textbf{FiCoCo}~\cite{Han2024:FiCoCo} is a two-stage compression method that merges tokens in ViT using \texttt{[CLS]} and patch-patch attention, then further compresses in LLM using text-visual attention. It suffers from \texttt{[CLS]} dependency and lacks Flash Attention compatibility.

    \item \textbf{FasterVLM}~\cite{Zhang:FasterVLM} is another pre-LLM compression method that relies on \texttt{[CLS]} token attention weights to retain informative visual tokens. It also faces compatibility issues with \texttt{[CLS]}-free VideoLLMs and Flash Attention integration in ViT.
    
    \item \textbf{DyCoke}~\cite{tao2024dycoke} is a two-stage VideoLLM-specific method that first prunes similar tokens along the temporal dimension and then uses attention weights in the LLM to compress the less attended visual tokens in the KV cache. Due to its reliance on dividing frame sets into parts and compressing them through similarity calculations, similar to ToMe~\cite{Bolya:ToMe}, it cannot achieve aggressive token compression in one go. While its token compression stage is compatible with Flash Attention~\cite{Dao2022:FlashAttention}, its KV cache compression requires explicit attention weights and thus remains incompatible with efficient attention operators.

\end{itemize}

\section{Sensitivity Analysis}
\label{sec:appendix/hyper-para}

\begin{table}[!t]
  \centering
  \setlength{\tabcolsep}{0.3pt}
  \scalebox{0.8}{
    \begin{tabular}{lcccccc}
    \multirow{2}{*}{\textbf{Metrics}} &  \multirow{2}{*}{\textbf{MVBench}}  &  \multicolumn{4}{c}{\textbf{VideoMME}} & \multirow{2}{*}{\textbf{Avg.}} \\
        & & \textbf{Overall} & \textbf{Short} & \textbf{Medium} & \textbf{Long}  & \\
    \shline
    \textcolor[rgb]{ .502,  .502,  .502}{Vanilla} & 
    \textcolor[rgb]{ .502,  .502,  .502}{56.9} &
    \textcolor[rgb]{ .502,  .502,  .502}{58.6} & 
    \textcolor[rgb]{ .502,  .502,  .502}{70.3} &
    \textcolor[rgb]{ .502,  .502,  .502}{56.6} &
    \textcolor[rgb]{ .502,  .502,  .502}{48.8} &
    \textcolor[rgb]{ .502,  .502,  .502}{100.0}
    \\
    \hline
    $u^\mathrm{frame}_{t,m}$ & 56.8  & 57.9  & 68.8  & 56.9  & 48.1 & 98.8 \\
    $u^\mathrm{video}_{t,m}$  & 56.8  &  58.3 & 69.3  &  56.1 & 49.3 & 99.3 \\
    \hline
    \multicolumn{7}{l}{\textit{Combination}} \\
    \textbf{$u^\mathrm{frame}_{t,m} + u^\mathrm{video}_{t,m}$} & 57.2 & 58.6 & 69.8 & 56.4 & 49.4 & 100.3 \\

    \textbf{$u^\mathrm{frame}_{t,m} + 2u^\mathrm{video}_{t,m}$} & 56.1 & 58.4 & 69.7 & 56.4 & 49.0 & 99.5 \\
    \textbf{$2u^\mathrm{frame}_{t,m} + u^\mathrm{video}_{t,m}$} & 56.9 & 58.6 & 69.7 & 56.8 & 49.3 & 100.0 \\

    \end{tabular}%
    }
  \vspace{-2mm}
  \caption{\textbf{Effects of balancing hyper-parameters between $u^\mathrm{frame}_{t,m}$ and $u^\mathrm{video}_{t,m}$ on VidCom$^2$ performance.}}
  \label{tab:hyper-para}%
  \vspace{-5mm}
\end{table}%

% \begin{table}[!t]
%   \centering
%   \setlength{\tabcolsep}{0.5pt}
%   \scalebox{0.76}{
%     \begin{tabular}{lccccccc}
%     \multirow{2}{*}{\textbf{Metrics}} & \multirow{2}{*}{\textbf{MVBench}}  &  \multirow{2}{*}{\textbf{MLVU}}  &  \multicolumn{4}{c}{\textbf{VideoMME}} & \multirow{2}{*}{\textbf{Avg.}} \\
%         & & & \textbf{Overall} & \textbf{Short} & \textbf{Medium} & \textbf{Long}  & \\
%     \shline
%     \textcolor[rgb]{ .502,  .502,  .502}{Vanilla} & \textcolor[rgb]{ .502,  .502,  .502}{56.9} &
%     \textcolor[rgb]{ .502,  .502,  .502}{63.0} &
%     \textcolor[rgb]{ .502,  .502,  .502}{58.6} & 
%     \textcolor[rgb]{ .502,  .502,  .502}{70.3} &
%     \textcolor[rgb]{ .502,  .502,  .502}{56.6} &
%     \textcolor[rgb]{ .502,  .502,  .502}{48.8} &
%     \textcolor[rgb]{ .502,  .502,  .502}{100.0}
%     \\
%     \hline
%     Uniform &   & 61.9  &  57.9 & 68.8  & 56.9  & 48.1 &   \\
%     \hline
%     \multicolumn{7}{l}{\textit{Frame Compression Adjustment}} \\
%     $\max{u^\mathrm{video}_{t,m}}$ &   & 62.1  & 58.1  &  68.4 & 56.7  & 49.3 &  \\
%     \rowcolor[rgb]{ .949,  .949,  .949}
%     $\overline{u^\mathrm{video}_{t,m}}$ &   & 62.3  &  58.2 & 69.1  & 55.9  & 49.6 &  \\
%     \end{tabular}%
%     }
%   \vspace{-2mm}
%   \caption{\textbf{Effects of different compression adjustment.} ``Uniform'': fixed $R=25\%$. ``$\max u^\mathrm{video}_{t,m}$'' and ``$\overline{u^\mathrm{video}_{t,m}}$'' denote frame uniqueness score $u_t$ of frame $t$ computed by maximum and average operations of $u^\mathrm{video}_{t,m}$.}
%   \label{tab:ablation_1_ours}%
%   \vspace{-4mm}
% \end{table}%

Table~\ref{tab:hyper-para} further explores the hyper-parameter that balances the influence of $u^\mathrm{frame}_{t,m}$ and $u^\mathrm{video}_{t,m}$ on $u_{t,m}$ in our VidCom$^2$ method. We observe that our method is not particularly sensitive to the balancing coefficient, as different degrees of balancing result in minimal performance differences. However, all balanced configurations outperform using either $u^\mathrm{frame}_{t,m}$ or $u^\mathrm{video}_{t,m}$ alone. This suggests that when performing token compression in VideoLLMs, it is crucial to consider the uniqueness of each token both within its frame and across the entire video to preserve more distinctive visual information. Notably, we find that $u_{t,m} = u^\mathrm{frame}_{t,m}+u^\mathrm{video}_{t,m}$ yields the best performance, indicating that $u^\mathrm{frame}_{t,m}$ and $u^\mathrm{video}_{t,m}$ are equally important. Therefore, we adopt $u_{t,m} = u^\mathrm{frame}_{t,m}+u^\mathrm{video}_{t,m}$ as our default configuration.

\begin{algorithm}[t]
\caption{VidCom$^2$: Plug-and-Play Token Compression for VideoLLMs}
\label{alg:vidcom2}
\begin{algorithmic}[1]
\Require  
Video tokens $\mathbf{X}^v = \{\mathbf{x}_{t,m}^v\}_{t=1,m=1}^{T,M}$,  
Preset retention ratio $R\in(0,1]$,  
Temperature $\tau>0$,  
Stability epsilon $\epsilon>0$
\Ensure Compressed tokens $\{\hat{\mathbf{X}}_t^v\}_{t=1}^T$

\State \textbf{Stage 1: Frame Compression Adjustment}  
\State // 1. Compute global summary  
\State $\displaystyle \mathbf{g}_v \leftarrow \tfrac{1}{T\cdot M}\sum_{t=1}^T\sum_{m=1}^M \mathbf{x}_{t,m}^v$
\State // 2. Token–video similarity  
\For{$t=1\to T,\ m=1\to M$}
  \State $s^\mathrm{video}_{t,m}\leftarrow \dfrac{\mathbf{x}_{t,m}^v\cdot\mathbf{g}_v}{\|\mathbf{x}_{t,m}^v\|\|\mathbf{g}_v\|}$
  \State $u_{t,m}^\text{video}\leftarrow -\,s^\mathrm{video}_{t,m}$
\EndFor
    \State // 3. Frame uniqueness  
\For{$t=1\to T$}
  \State $u_t\leftarrow \frac{1}{M}\sum_{m=1}^{M} u^\mathrm{video}_{t,m}$
\EndFor
    \State // 4. Normalize \& weigh  
\For{$t=1\to T$}
  \State $\tilde u_t\leftarrow (u_t - \max_k u_k)/\tau$
  \State $\sigma_t\leftarrow \exp(\tilde u_t)\,/\,(\sum_{k=1}^T\exp(\tilde u_k)+\epsilon)$
  \State $r_t\leftarrow R\bigl(1 + \sigma_t - \tfrac1T\bigr)$  
\EndFor

\State \textbf{Stage 2: Adaptive Token Compression} 
\For{$t=1\to T$}
  \State // 1. Frame-level token uniqueness  
  \State $\mathbf{g}_{f,t}\leftarrow \tfrac{1}{M}\sum_{m=1}^M \mathbf{x}_{t,m}^v$
  \For{$m=1\to M$}
    \State $s_{t,m}^\text{frame}\leftarrow \,\dfrac{\mathbf{x}_{t,m}^v\!\cdot\!\mathbf{g}_{f,t}}{\|\mathbf{x}_{t,m}^v\|\|\mathbf{g}_{f,t}\|}$
    \State $u_{t,m}^\text{frame} \leftarrow -s_{t,m}^\text{frame}$
  \EndFor
  \State // 2. Combine video \& frame uniqueness  
  \For{$m=1\to M$}
    \State $u_{t,m}\leftarrow u_{t,m}^\text{video} + u_{t,m}^\text{frame}$
  \EndFor
  \State // 3. Top-$k$ selection  
  \State $k_t\leftarrow \lceil r_t \times M\rceil$
  \State $\hat{\mathbf{X}}_t^v \leftarrow \text{TopK}(\{\mathbf{x}_{t,m}^v\}, \{u_{t,m}\}, k_t)$
\EndFor

\State \Return $\{\hat{\mathbf{X}}_t^v\}_{t=1}^T$
\end{algorithmic}
\end{algorithm}

\section{Algorithm Details of VidCom$^2$}
\label{sec:appendix/algorithm}

Algorithm~\ref{alg:vidcom2} present the algorithm workflow of our VidCom$^2$. This algorithm details the step-by-step process of our token compression framework, illustrating how VidCom$^2$ dynamically adjusts compression intensity based on frame uniqueness and preserves the most distinctive tokens both within each frame and across the entire video sequence.

\section{More Visualization of Frame Uniqueness}
\label{sec:appendix/more_vis}

\begin{figure*}[!t]
  \centering
   \includegraphics[width=\linewidth]{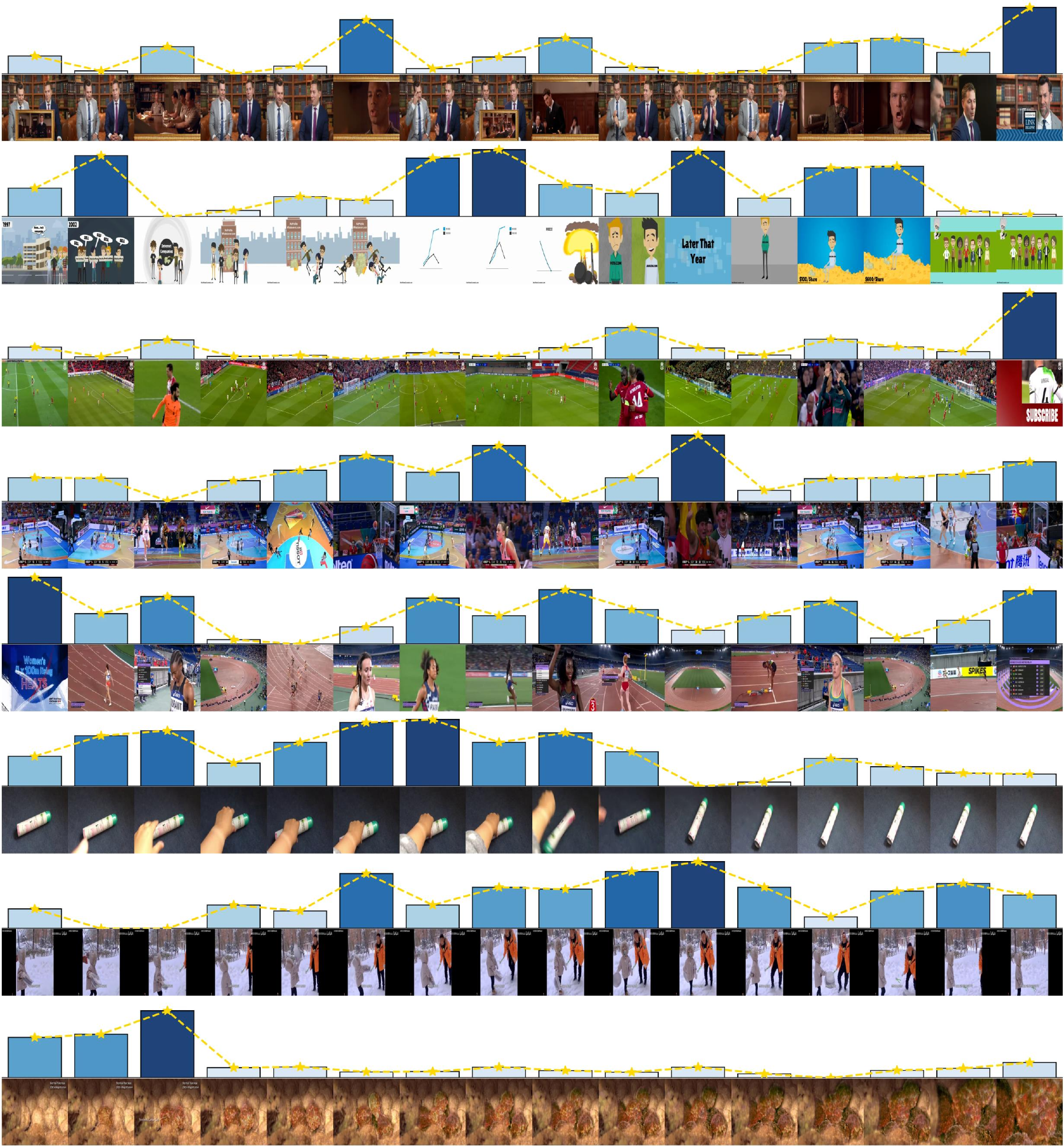}
   \vspace{-7mm}
    \caption{\textbf{More visualization of frame uniqueness quantified by our VidCom$^2$.} In most cases, the frame uniqueness determined by VidCom$^2$ aligns well with human video perception.}   
   \label{fig:frame_uniqueness_appendix}
   \vspace{-4mm}
\end{figure*}

Figure~\ref{fig:frame_uniqueness_appendix} presents additional visualizations of frame distinctiveness as quantified by our VidCom$^2$. These cases cover a diverse range of scenarios, including everyday life situations, sports activities, dynamic scenes, and scientific domains. The visualizations demonstrate that VidCom$^2$ effectively quantifies frame uniqueness across these varied contexts, consistently allocating more token budget to distinctive frames. This approach ensures the preservation of more visually unique information across diverse scenarios, which is crucial for accurate video understanding by VideoLLMs.

\end{document}